\documentclass{article}
\usepackage{spconf,amsmath,graphicx}
\usepackage{graphicx}
\usepackage{amsmath,amssymb} 
\usepackage{algorithm}
\usepackage{algpseudocode}
\usepackage{amsmath}
\usepackage{amsmath} 
\usepackage{amsfonts,amssymb}
\usepackage{multirow, array}
\usepackage{subfigure}
\usepackage{multirow}
\usepackage{hyperref}
\usepackage{amssymb}
\usepackage[misc]{ifsym} 
\newenvironment{shrinkeq}[1]
{ \bgroup
\addtolength\abovedisplayshortskip{#1}
\addtolength\abovedisplayskip{#1}
\addtolength\belowdisplayshortskip{#1}
\addtolength\belowdisplayskip{#1}}
{\egroup\ignorespacesafterend}

\title{BLP - Boundary Likelihood Pinpointing Networks for Accurate Temporal Action Localization}
\name{Weijie Kong$^{1,2}$, Nannan Li$^1$, Shan Liu$^3$, Thomas Li$^4$, Ge Li$^{*1,2}$\thanks{*Corresponding Author. This project was supported by 
Shenzhen Key Laboratory for Intelligent Multimedia and Virtual Reality (ZDSYS201703031405467), 
Shenzhen Municipal Science and Technology Program under Grant JCYJ20170818141146428,
National Engineering Laboratory for Video Technology - Shenzhen Division,
and Peng Cheng Laboratory. We would like to thank Jia-Xing Zhong and Tao Zhang for their helpful comments and discussion.}}
\address{$^1$School of Electronic and Computer Engineering, Peking University, Shenzhen, China.\\
$^2$ Peng Cheng Laboratory, Shenzhen, China.\\
$^3$ Media Lab, Tencent, Shenzhen, China.\\
$^4$ Advanced Institute of Information Technology, Peking University, Hangzhou, China. \\
}
\begin{document}
\ninept
\maketitle
\begin{abstract}
Despite tremendous progress achieved in temporal action detection, state-of-the-art methods still suffer from the sharp performance deterioration when localizing the starting and ending temporal action boundaries. Although most methods apply boundary regression paradigm to tackle this problem, we argue that the direct regression lacks detailed enough information to yield accurate temporal boundaries. In this paper, we propose a novel Boundary Likelihood Pinpointing (BLP) network to alleviate this deficiency of boundary regression and improve the localization accuracy. Given a loosely localized search interval that contains an action instance, BLP casts the problem of localizing temporal boundaries as that of assigning probabilities on each equally divided unit of this interval. These generated probabilities provide useful information regarding the boundary location of the action inside this search interval. Based on these probabilities, we introduce a boundary pinpointing paradigm to pinpoint the accurate boundaries under a simple probabilistic framework. Compared with other C3D feature based detectors, extensive experiments demonstrate that BLP significantly improves the localization performance of recent state-of-the-art detectors, and achieves competitive detection mAP on both THUMOS' 14 and ActivityNet datasets, particularly when the evaluation tIoU is high.





\end{abstract}

\vspace{-1.0em}
\section{Introduction}
\label{sec:intro}

Recently, as an essential but challenging task in the large research scope of video analysis, temporal action detection in untrimmed videos has drawn tremendous attention from the research community \cite{shou2016scnn,shou2017cdc,zhao2017ssn,gao2017cascaded,dai2017tcn,xu2017r,lin2017sst,sstad_buch_bmvc17,Yang:2017wa,heilbron2017scc,Lea:2017fy}. Given a long untrimmed video consisting of multiple action instances and complex background contents, temporal action detection aims at solving two problems: (1) recognizing the action categories of actions contained in the video; (2) localizing temporal intervals (starting and ending boundaries) where actions of interest occur. Currently, temporal action detection has been applied to multiple practical applications, such as video surveillance, human-robot interaction and intelligent home care. 

 For temporal action detection, how to accurately localize the starting and ending boundaries of a complex action instance is a challenging problem, since an action instance can happen at arbitrary temporal location with uncertain duration in a video of diverse length. As addressed in \cite{alwassel_2018_detad}, \emph{the localization error is the most common and the most impactful error that hampers the detection performance of existing state-of-the-art approaches.} Preferentially fixing localization errors can significantly boost the detection average-mAP.
Therefore, to achieve high temporal localization accuracy, most recently detection methods \cite{zhao2017ssn,gao2017cascaded,xu2017r,lin2017sst,lin2017temporal,gao2017turn,chao2018rethinking} apply boundary regression paradigm to refine the boundaries given a proposal. However, we argue that trying to directly regress the action boundary temporally constitutes a difficult learning task and hardly yield accurate enough boundaries. 
\begin{figure}
\centering
\includegraphics[width=8.5cm] {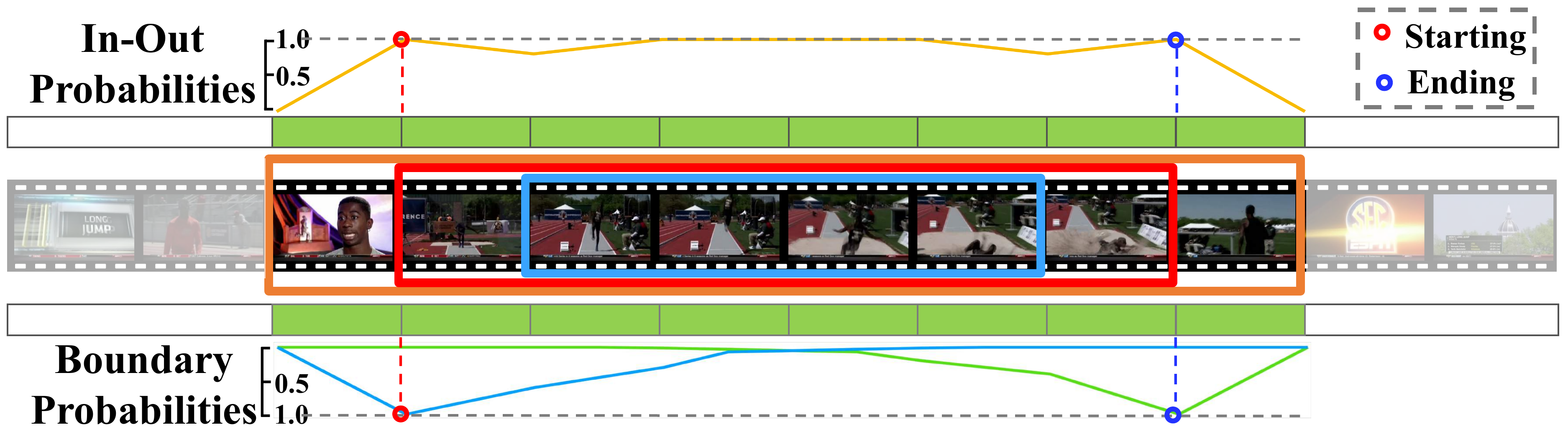}
\caption{\textbf{Localization process of boundary pinpointing paradigm.} A search interval is extended from an action proposal by a factor $\gamma$, and equally divided into $M$ units. To localize the precise temporal boundary, we assign \emph{in-out} or \emph{boundary} probabilities to each equally divided unit of the search interval for being on the inside of an action temporal boundary or being the starting or ending boundary of the action instance. By maximizing the likelihood of each probability, we pinpoint the predicted temporal boundaries (red box).}
\label{fig:illustration}
\end{figure}


\emph{To alleviate this deficiency and focus on the need for improving the localization accuracy of current detection methods, we propose a novel Boundary Likelihood Pinpointing (BLP) network.}
The \emph{main contribution} of BLP is that we cast the problem of localizing temporal boundaries as that of assigning probabilities on each equally divided unit of a search interval. Specifically, instead of using boundary regression, we propose a \emph{boundary pinpointing} paradigm to perform accurate temporal action localization, which is implemented with three steps (see Fig. \ref{fig:illustration}). First, given a loosely localized action proposal within a video, we obtain a larger search interval via extending the proposal boundaries by a factor $\gamma$ and equally divide it to $M$ units. Second, we assign one or more discrete probabilities to each unit indicating whether the unit is inside of the temporal span of action ground truth or being the starting or ending boundary of the action instance. Finally, we pinpoint the boundaries by simply maximizing the likelihood for estimating the optimal boundaries under these probabilities. Since these probabilities provide far more detailed and useful boundary information, they would encourage the model to yield more accurate boundaries than the regression models, that just predict 2 temporal boundary coordinates. We evaluate BLP model on two challenging datasets: THUMOS'14 \cite{THUMOS14} and ActivityNet \cite{caba2015activitynet}. Extensive experiments demonstrate that BLP model can obtain detection results with more precise boundaries than direct regression.  Integrating our BLP model with existing action classifier into detection framework leads to competitive detection mAP on both datasets, especially when the evaluation tIoU is high. Specifically, our detection framework achieves 34.5\% (tIoU = 0.7) and 15.5$\times$ (tIoU = 0.95) relative gain over the mAP of the state-of-the-arts on THUMOS'14 and ActivityNet respectively.
\textbf{Relation to prior work.} Recently, an immense amount of deep models \cite{simonyan2014two,tran2015learning,Wang:2016fa,carreira2017quo,wu2018compressed} have been proposed in action recognition, among which the Two-Stream \cite{simonyan2014two} and C3D \cite{tran2015learning} models are deployed in most existing methods. Due to the explosive growth of untrimmed video data, another challenging task called temporal action detection has been put to the center of attention. Currently, many approaches \cite{xu2017r,gao2017cascaded,zhao2017ssn,lin2017sst,gao2017turn,chao2018rethinking} adopt a ``detection by classification'' framework, in which boundary regression has been widely employed for adjusting the temporal boundaries and boosting the localization accuracy. 
 Differently from the aforementioned work, we propose the \emph{boundary pinpointing} paradigm. This paradigm estimates the optimal boundaries by maximizing likelihood under predefined probabilities. where these probabilities can provide more useful information than dirct regression. Our idea stems from a novel \emph{object localization methodology} called LocNet \cite{gidaris2016locnet}, which proposed to revise the horizontal and vertical object boundaries of the given proposal using border probabilities. 
 Inspired by this work, BSN \cite{BSN2018arXiv} also adopted similar boundary probabilities for \emph{temporal action proposal}. However, BSN generates proposal boundaries by simply selecting temporal locations with high starting and ending probabilities separately. Our method localizes temporal boundaries using maximizing likelihood estimation under these probabilities, which can provide a more accurate measure of confidence for delimiting the boundaries on any point in time.

\vspace{-1.5em}
\section{Proposed Method}
\label{sec:approach}

\vspace{-0.5em}
\subsection{Temporal Action Detection Pipeline}
\label{sec:overview}

To begin with, we provide a brief overview of the temporal action detection pipeline. \emph{Our detection pipeline contains two major modules: an action classification and localization network}. 

Formally, an action proposal is represented as $\psi=(B_s,B_e)$, where $B_s$ and $B_e$ are the starting ($s$) and ending ($e$) boundary coordinates of the segment $\psi$ separately. Given a set of action proposals $\mathbf{\Psi}={\lbrace \psi_n \rbrace}^{N_p}_{n=1}$ generated by either sliding temporal window or other temporal action proposal methods \cite{escorcia2016daps,gao2017turn,BSN2018arXiv}, the \emph{action classification network} anticipates action categories by predicting a set of classification scores ${\lbrace\lbrace s_{nj}|c_j \rbrace}^{N_c}_{j=1}\rbrace^{N_p}_{n=1}$. The score $s_{nj}$ represents how likely the n-th temporal proposal is recognized to be the j-th action category $c_j$. Meanwhile, for each loosely localized proposal, the \emph{action localization network} localizes boundaries where actions start and end temporally. It generates a new set of action segments that have more compact boundaries enclosing the actions inside the proposal. To eliminate the redundant segments, an extra Non-Maximum-Suppression (NMS) \cite{softnms} operation is applied to obtain the final segments with accurate boundaries $\mathbf{\Psi^{'}}={\lbrace \psi_n^{'}=(B_{ns}^{'},B_{ne}^{'}) \rbrace}^{N_d}_{n=1}, (N_d \leq N_p)$. Details on the localization process will be discussed in Sec. \ref{sec:blp}. $N_c$, $N_d$ and $N_p$ are the number of action categories, final results and proposals respectively.

\vspace{-1.0em}
\subsection{Boundary Likelihood Pinpointing Network}
\label{sec:blp}
\emph{The purpose of our work is to improve the localization accuracy of the detection pipeline.} Currently, most existing detection methods \cite{zhao2017ssn,gao2017cascaded,xu2017r,lin2017sst,lin2017temporal,gao2017turn,chao2018rethinking} accomplish this by directly regressing two boundary coordinates, which lacks detailed enough information to yield accurate boundaries. Thus, we propose a novel \textbf{Boundary Likelihood Pinpointing (BLP)} network as our localization network.


BLP accepts selected proposal segments and outputs conditional probabilities indicating the boundary location. Given a proposal segment $\psi=(B_s,B_e)$, BLP first extends it by a factor $\gamma$ to create a search interval $I=(\frac{(1+\lambda)B_s+(1-\lambda)B_e}{2},\frac{(1-\lambda)B_s+(1+\lambda)B_e}{2}) $ and equally divided it into $M$ units. Then, BLP predicts one or more discrete probabilities for each unit to indicating whether the unit is inside of the temporal span of action ground truth or being the starting or ending boundary of the action instance. These probabilities provide more detailed information for precise boundary inference than direct boundary regression, which is detailed in Sec. \ref{sec:localization_model}. During inference, we propose a novel boundary pinpointing paradigm. Based on the probabilities generated by BLP, we can pinpoint the action boundaries by simply maximizing the likelihood for estimating the optimal boundaries. This paradigm is detailed in Sec. \ref{sec:boundary_pinpointing}.




\vspace{-1.0em}
\subsubsection{Boundary Likelihood Predictions}
\label{sec:localization_model}
 For each unit $i$ within a search interval $I$, BLP predicts one or more conditional probabilities $p^{I,c}={\lbrace p(i|I,c)\rbrace}^M_{i=1}$ corresponding to a specific category $c$. Here we design two types of probabilities.

\textbf{In-Out probabilities:} We define the \emph{in-out} probabilities $p_{io}={\lbrace p_{io}(i) \rbrace}^M_{i=1}$ to represent the likelihood of unit $i$ being inside the temporal span of an action instance of category $c$. Ideally, given a ground truth segment $\psi_{gt}=(B^{gt}_s, B^{gt}_e)$, the \emph{in-out} probabilities $p_{io}$ should be equal to the following target probabilities $T={\lbrace T_{io} \rbrace}$.
\begin{shrinkeq}{-1.5ex}
\begin{displaymath}
\forall i \in \lbrace1,...,M\rbrace,
	T_{io}(i)=
	\begin{cases}
	1,& \text{if  unit $i\in [B^{gt}_s, B^{gt}_e$]} \\
	0,& \text{otherwise}
	\end{cases}
\end{displaymath}
\end{shrinkeq}

\textbf{Boundary probabilities:} $p_s=\lbrace{p_s(i)}\rbrace^M_{i=1}$ and  $p_e=\lbrace{p_e(i)}\rbrace^M_{i=1}$ represent two independent probabilities of unit $i$ being the starting and ending boundaries of an action instance for category $c$. Given a ground truth $\psi_{gt}$, the output \emph{boundary} probabilities $p_{bd}={p_s, p_e}$ should ideally equal to target probabilities $T={\lbrace T_{s}, T_{e} \rbrace}$, where  $l\in\lbrace s, e\rbrace$.
\begin{shrinkeq}{-1.5ex}
\begin{displaymath}
\forall i \in \lbrace1,...,M\rbrace,
	T_{l}(i)=
	\begin{cases}
	1,& \text{if $B^{gt} \in$ unit $i$}  \\
	0,& \text{otherwise}
	\end{cases}
\end{displaymath}
\end{shrinkeq}



\vspace{-1.5em}
\subsubsection{Inference by Boundary Pinpointing}
\label{sec:boundary_pinpointing}
Given aforementioned probabilities of $I$, we propose a novel boundary pinpointing paradigm to inference the temporal boundaries $\psi^{'}=(B_{s}^{'},B_{e}^{'})$ of the action inside $I$. This process is implemented by adopting one of the following two BLP localization models.


\textbf{In-Out localization model:} Maximizes the likelihood of \emph{in-out} elements of temporal boundary $\psi^{'}$:
\begin{shrinkeq}{-1.5ex}
\begin{equation}
L_{in-out}(\psi^{'}) = \prod_{i\in\lbrace B_s,...,B_e\rbrace} p_{io}(i)\prod_{i\notin\lbrace B_s,...,B_e\rbrace} \widetilde{p_{io}}(i),
\end{equation}
\end{shrinkeq}
where $\widetilde{p_{io}}(i)=1-p_{io}(i)$. The first term in the right hand of the equation represents each unit of $\psi^{'}$ to be inside a ground truth interval and the second term represents the likelihood of units that are not part of $\psi^{'}$ to be outside a ground truth interval.

\textbf{Boundary localization model:} Maximizes the likelihood of \emph{boundary} elements of boundary $\psi^{'}$:

\begin{shrinkeq}{-1.5ex}
\begin{equation}
L_{boundary}(\psi^{'}) = p_s(B_s^{'})\cdot p_e(B_e^{'}).
\end{equation}
\end{shrinkeq}



\vspace{-0.5em}
\subsection{Action Detection Network Architecture} 
\begin{figure}[t]
\setlength{\belowcaptionskip}{-0.5cm}
\centering
\includegraphics[width=8.4cm]{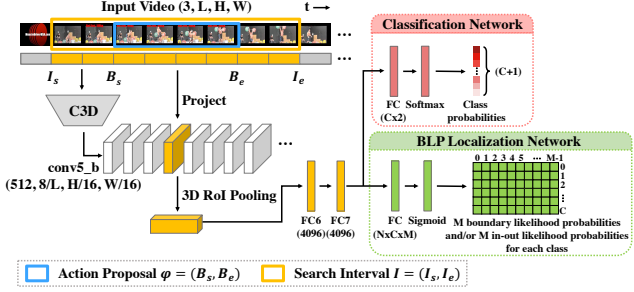}
\caption{Temporal action detection network architecture with BLP localization model.}
\label{fig:architecture}
\end{figure}

The architecture of the detection network is shown in Fig. \ref{fig:architecture}. Given a video sequence $\mathbf{V} \in \mathbb{R}^{3 \times L \times H \times W}$ consists of $L$ frames  and a set of action proposals $\mathbf{\Psi}={\lbrace \psi_n \rbrace}^{N_p}_{n=1}$, the network outputs category-specific action segments with accurate temporal boundaries. 


\textbf{BLP localization network architecture.} BLP network aims to predict aforementioned \emph{in-out} or \emph{boundary} probabilities for each proposal.
To begin with, a deep shared C3D model \cite{tran2015learning} is utilized to process the input video $\mathbf{V}$ to extract rich spatio-temporal feature hierarchies, and outputs a shared feature map $\mathbf{F_{conv5b}}\in \mathbb{R}^{512 \times \frac{L}{8} \times \frac{H}{16} \times \frac{W}{16}}$. 
Then, given a search interval $I$ extended by an action proposal, we map it on $\mathbf{F_{conv5b}}$ and use a 3D RoI pooling layer \cite{xu2017r} to extract fixed-size feature maps (of size $512 \times 1 \times 4 \times 4$) from activation that inside $I$. The resulting feature maps can be fed forward into two fully connected (fc) layers of C3D and an extra fc layer to yields a 1-dimension feature vector with length $N \times M \times C$, where $N=1, 2$ for \emph{in-out} and \emph{boundary} possibilities respectively, $M$ is the number of divided units of $I$ and $C$ is the number of action categories.  
Finally, in order to output the category-specific conditional probabilities, the 1-dimension feature vector is reshaped and fed into a sigmoid layer to obtain the final conditional probability matrix with dimension $N \times M \times C$.

\textbf{Action classification network architecture.} For a given proposal, action classification network anticipates action categories by predicting a set of softmax scores for $(C+1)$ categories (including ``background''). To this end, the fc7 features are fed into another fc layer and an extra softmax layer to output $(C+1)$ class probabilities. 



\vspace{-2.0em}
\subsection{Optimization} 
We train the detection network by optimizing classification and localization networks jointly. The multi-task objective function is:
\begin{shrinkeq}{-1.5ex}
\begin{equation}
\label{equ:loss}
\begin{split}
\mathcal{L}oss = \frac{1}{N_{cls}}\sum_i{\mathcal{L}_{cls}(\theta_1|a_i, a_i^*)} 
+\frac{\lambda}{N_{loc}}\sum_j{\mathcal{L}_{loc}(\theta_2|p_{(\cdot)j}, T_{(\cdot)j}, c_j)}
\end{split}
\end{equation}
\end{shrinkeq}
where $N_{cls}$ and $N_{loc}$ stand for batch size and number of proposal segments, respectively, and $\lambda$ is the trade-off parameter and set empirically. 
The $i$ and $j$ are the indexes for action proposals, and $\theta_.$ are the network parameters. For the classification network, $\mathcal{L}_{cls}$ is a standard multi-class cross-entropy loss, where $a_i$ and $a^*_i$ are the predicted class probability and the ground truth, respectively; while for the localization network, $L_{loc}$ adopts a binary logistic regression loss conditioned on a specific class $c$, where $p_{(.)j}=\{p_{(io)j}, p_{(db)j} \}$ represent evaluation probabilities of $in$-$out$ or $boundary$ for each segment, and $T_{(\cdot)j}=\{T_{(io)j},T_{(bd)j} \}$ are the corresponding target probabilities. Specifically, in the case of \emph{in-out}, the loss $\mathcal{L}_{loc}$ is given by:
\begin{shrinkeq}{-1.5ex}
\begin{equation}
\mathcal{L}_{loc} = \sum^M_{j=1}T_{(io)j}log(p_{(io)j})+\widetilde{T}_{(io)j}log(\widetilde{p}_{(io)j}),
\end{equation}
\end{shrinkeq}
for the \emph{boundary} case, it is:
\begin{shrinkeq}{-1.5ex}
\begin{equation}
\label{equ:bd_loss}
\mathcal{L}_{loc} = \sum_{bd\in{\lbrace s,e \rbrace}} \sum^M_{j=1} \beta^+ T_{(bd)j}log(p_{(bd)j})+ \beta^- \widetilde{T}_{(bd)j}log(\widetilde{p}_{(bd)j}),
\end{equation}
\end{shrinkeq}
where $\widetilde{p}_{(.)j} = 1-p_{(.)j}$, and $\widetilde{T}_{(\cdot)j}=1-T_{(.)j}$. In equation (\ref{equ:bd_loss}), we adapt the trade-off parameters $\beta^- = 0.5M/(M-1)$ and $\beta^+=(M+1)\cdot\beta^-$ as in \cite{gidaris2016locnet} to balance the two terms of $boundary$ and \emph{non-boundary} elements.
  
\vspace{-1.0em}
\section{Experiments}
\label{sec:experiments}
In this section, we evaluate proposed BLP network on two prevailing datasets: THUMOS'14 \cite{THUMOS14} and ActivityNet \emph{v1.3} \cite{caba2015activitynet}. \textbf{Baseline Model:} We take R-C3D \cite{xu2017r} as our baseline, since it's a regression-based temporal action detection method.  
To detect actions, we integrate examined BLP localization model and R-C3D classification model into one holistic detection framework.
For a fair comparison, we train and test our detection network with the same classification network and the same proposal set generated by  R-C3D for all experiments. The whole BLP model is implemented on Caffe \cite{jia2014caffe}. 
 
\vspace{-1.0em}
\subsection{Datasets and Experimental Details}
\textbf{THUMOS'14.} THUMOS'14 contains 20 different sport activities, with 200 videos for training and 213 videos for testing. \textbf{Evaluation metrics.} We report the mean Average Precision (mAP) of each action category at tIoU thresholds with [0.1:0.1:0.7], and the mAP at tIoU=0.5 is used for the final comparison with other methods. \textbf{Implementation details.}  The weights of C3D model are pre-trained on Sport-1M and finetuned on UCF101. The $\lambda$ in loss function (\ref{equ:loss}) is set to be 20. Other implementation details are the same as in \cite{xu2017r}.

\textbf{ActivityNet.} ActivityNet \emph{v1.3} contains 19,994 videos with 200 classes and is divided into three sets: training, validation, testing with a ratio of 2:1:1. \textbf{Evaluation metrics.} We report the mAP at tIoU=$\lbrace$0.5, 0.75, 0.95$\rbrace$, and the average of mAPs with tIoU thresholds [0.5:0.05:0.95] is used for comparison. \textbf{Implementation details.} The C3D model is initialized with the pre-trained Sport-1M weights finetuned on ActivityNet training videos. We train the BLP with a learning rate fixed at $10^{-4}$ for first 10 epochs and decreased to $10^{-5}$ for the last 5 epochs. The $\lambda$ is set to be 250.


\vspace{-1.0em}

\subsection{Ablation Experiments}
In this section, we explore the best hyper-parameter settings for BLP.

\textbf{How many units should a search interval be divided into? } Given a video search interval, we divide it to $M$ units. To explore the influence of $M$, we examine three \emph{In-Out} models with $M=\lbrace 16, 32, 48\rbrace$ (the extension factor $\gamma=1.8$). As shown in Table \ref{tab:M}, the detection performance with \emph{In-Out} model achieves the best performance when $M=32$. We analyze that with finer resolution ($M=48$), each unit contains fewer features to determine whether the unit is inside an action of interest. Conversely, with coarse resolution ($M=16$), each unit spans a longer time interval, therefore the temporal boundary localization may be ambiguous and less precise. The same analysis can be applied to \emph{Boundary} models. \emph{As a result, we choose M = 32 for the following experiments.} 



\textbf{How long should a proposal be extended to?} A search interval is obtained by extending a temporal segment by a factor $\gamma$. Our intuitive assumption is that with larger $\gamma$, the BLP model will comprehend and leverage more surrounding temporal context.  To explore the impact of $\gamma$, we investigate six \emph{In-Out} and \emph{Boundary} models with $\gamma= \lbrace 1.0, 1.6, 1.8, 2.0, 2.4, 3.0 \rbrace $ ($M=32$). As shown in Table \ref{tab:lambda}, we observe that when $\gamma=2.0$, two models achieves the peak performance, while the worst performance occurs when no context is considered ($\gamma=1.0$). However, including redundant context ($\gamma>2.0$) also leads to the deterioration of performance.   \emph{Thus, we choose $\gamma$ = 2.0 for the following experiments.}
\vspace{-2.0em}
\begin{table}[h]
\caption{Ablation experiment results on hyper-parameter $M$ for \emph{In-Out} and \emph{Boundary} models ($\gamma=1.8$, \%mAP@tIou).}
\centering
\begin{tabular}{p{1.3cm}<{}| p{0.9cm}<{\centering} p{0.9cm}<{\centering} p{0.9cm}<{\centering} p{0.9cm}<{\centering} p{0.9cm}<{\centering}}
\hline
tIoU & 0.1           & 0.2           & 0.3           & 0.4           & 0.5           \\ \hline\hline
M=16 & 54.8          & 52.7          & 47.9          & 39.4          & 31.2          \\
M=32 & \textbf{54.9} & \textbf{52.9} & \textbf{48.5} & \textbf{40.3} & \textbf{31.6} \\ 
M=48 & 53.3          & 51.1          & 47.1          & 39.7          & 29.6          \\ \hline
\end{tabular}
\label{tab:M}
\end{table}

\vspace{-2.0em}
\begin{table}[h]
\caption{Ablation experiment results on hyper-parameter $\gamma$ for \emph{In-Out} model ($M=32$, \%mAP@tIou=0.5).}
\centering
\begin{tabular}{p{1.3cm}<{}| p{0.7cm}<{\centering} p{0.7cm}<{\centering} p{0.7cm}<{\centering} p{0.7cm}<{\centering} p{0.7cm}<{\centering} p{0.7cm}<{\centering}}
\hline
$\gamma$ & 1.0 & 1.6 & 1.8 & 2.0     & 2.4           & 3.0\\ \hline
\emph{In-Out} & 30.5         & 31.3          & 31.6          & \textbf{32.1}& 31.8          & 31.7  \\ \hline
\emph{Boundary} & 29.3         & 32.4          & 32.2          & \textbf{32.5}& 31.9          & 31.9  \\ \hline
\end{tabular}
\label{tab:lambda}
\end{table}

\vspace{-2.0em}
\subsection{Action Localization Effectiveness Analysis} 
\begin{figure}[h]
  \centering
	
  \subfigure{
    \label{fig:subfig:recall1} 
    \includegraphics[width=3.9cm]{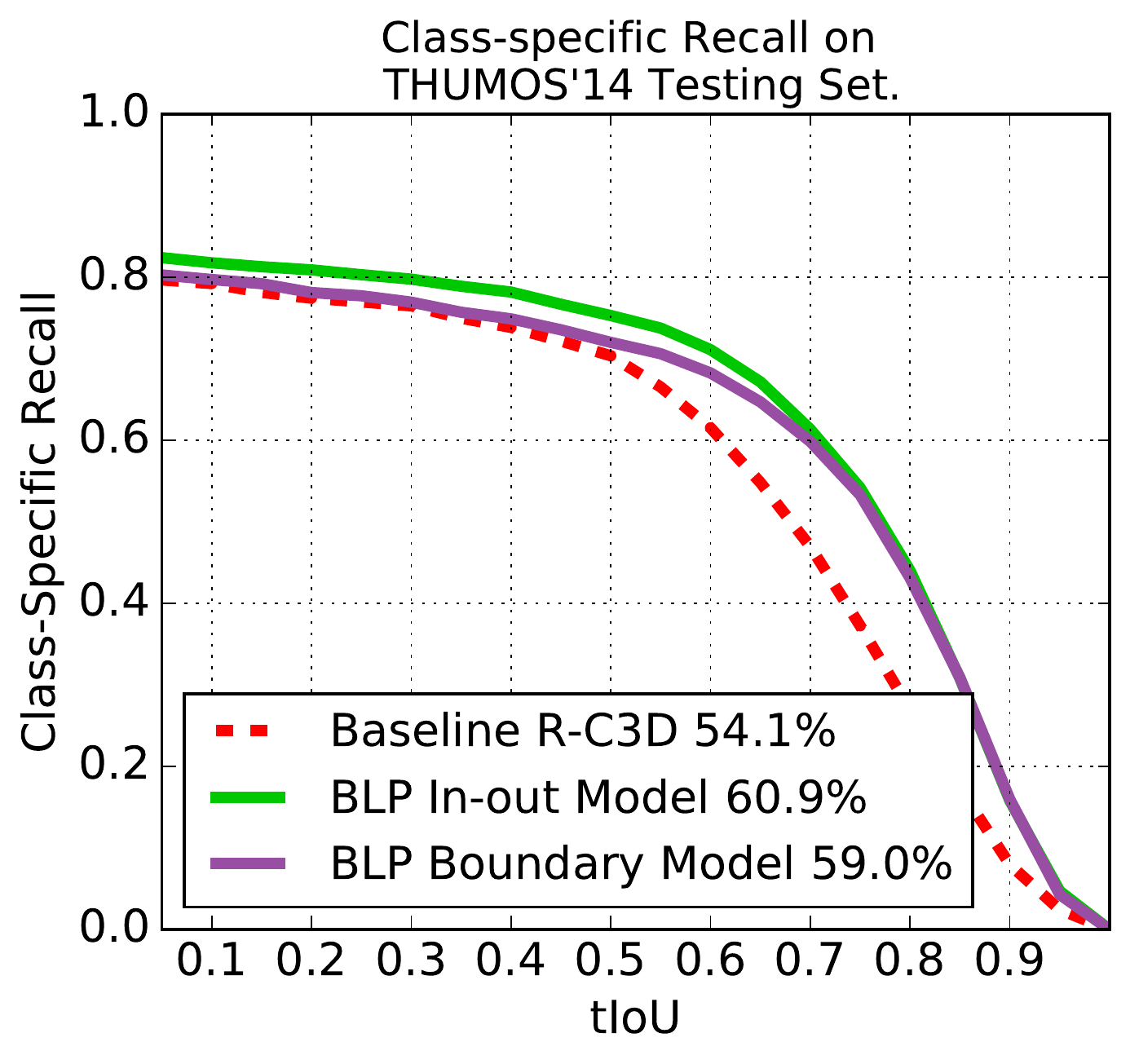}
    }
  \subfigure{
  	\label{fig:subfig:recall3} 
  	\includegraphics[width=3.9cm]{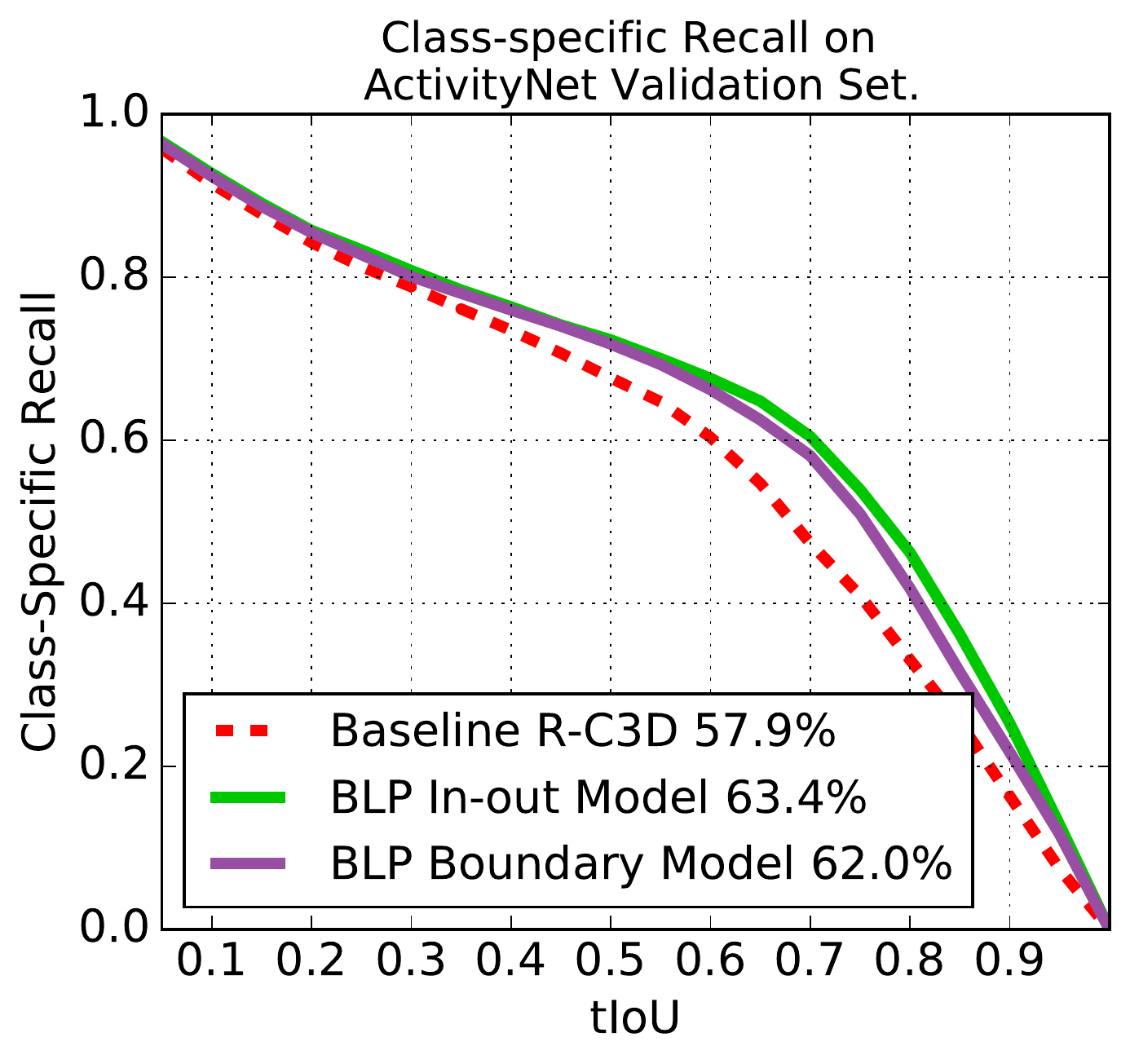}
  }
    
  \vspace{-1.0em}
  \caption{ Localization performance comparison on THUMOS' 14 and ActivityNet under matric: Class-Specific Recall@tIoU. For comparison, the average recall (AR) are reported in the legend.} 
  \label{fig:recall_loc} 
\end{figure}

In this section, we compare the localization performance of proposed \emph{In-Out} and \emph{Boundary} models with the regression-based model R-C3D on THUMOS'14 testing set and ActivityNet validation set.
As shown in Fig. \ref{fig:recall_loc}, to evaluate the localization performance of examined model, we report the class-specific recall (averaging per class recalls) as a function of the tIoU thresholds with [0.05:0.05:1.0] for the final detection results generated by the corresponding detection pipeline. We also report the average recall (AR) for each model in the legend. The higher AR indicates the model can yield the more accurate temporal boundaries. 
\emph{Fig. \ref{fig:recall_loc} shows that two proposed models achieve remarkably higher recall than the baseline, and surpass the AR of the baseline by average 5.9\% and 4.8\% on both datasets respectively. }
We argue that \emph{in-out} and \emph{boundary} probabilities help the BLP model to yield more accurate boundaries to have larger overlap with ground truth instances. 
This demonstrates the effectiveness and superior localization performance of BLP model. 
\vspace{-1.0em}

\vspace{-2.0em}
\subsection{Action Detection Performance Analysis}
 The detection performance is highly related to the choice of feature extractor. Since the Two-Stream features \cite{simonyan2014two,Wang:2016fa} or other improved 3D ConvNet features \cite{carreira2017quo} are more discriminative and superior than the vanilla C3D features deployed in our model, \emph{here we only compare with the state-of-the-art methods that adopt vanilla C3D as their feature extractors for a fair comparison.} 

\textbf{THUMOS'14.} The comparison results on THUMOS'14 testing set are summarized in Table \ref{tab:thumos-sota}. We can observe that: 
(1) The detection framework with proposed \emph{In-Out} and \emph{Boundary} localization models outperform the baseline R-C3D \cite{xu2017r} by \textbf{3.2\%} and \textbf{3.6\%} respectively, which demonstrates that our proposed boundary pinpointing paradigm can truly boost the localization performance and yield much more accurate temporal boundaries than boundary regression.
(2) When it comes to the other well-developed regression-based detection methods \cite{gao2017turn,gao2017cascaded}, the detection performance of our emerging probability-based localization method is remarkably superior.
(3) Our detector shows superior mAP over state-of-the-art methods SS-TAD \cite{sstad_buch_bmvc17} across a wide range of tIoU thresholds. Especially when tIoU = 0.7, we outperform SS-TAD by 35.4 relatively. These results confirm that our well-designed probabilities can provide more useful boundary information for accurate localization.
\vspace{-1.0em}
\begin{table}[h]
\centering
\caption{Temporal action detection results on THUMOS'14 testing set (\%mAP@tIoU). Here we only list  C3D feature based methods.}
\ninept
\begin{tabular}{p{2.8cm}<{\raggedright}||p{0.3cm}<{\centering} p{0.3cm}<{\centering} p{0.3cm}<{\centering} p{0.3cm}<{\centering} p{0.3cm}<{\centering} p{0.3cm}<{\centering} p{0.3cm}<{\centering}}

\hline
Detection Method       & 0.7 & 0.6 & 0.5 & 0.4 & 0.3 & 0.2 & 0.1 \\ \hline \hline 
%
SCNN \cite{shou2016scnn} & 5.3 & 10.3 & 19.0 & 28.7 & 36.3 & 43.5 & 47.7 \\
CBR-C3D \cite{gao2017cascaded} & 7.9 & 13.8 & 22.7 & 30.1 & 37.7 & 44.3 & 48.2    \\
CDC \cite{shou2017cdc}          & 7.9 & 13.1 & 23.3 & 29.4 &40.1 & - & -    \\
TURN + S-CNN \cite{gao2017turn} & - & - & 25.6 & 34.9 & 44.1 & 50.9 & 54.0\\
SS-TAD \cite{sstad_buch_bmvc17}   &9.6 & - & 29.2 & - &45.7 & - & -   \\ 
R-C3D (Baseline) \cite{xu2017r}&- & - & 28.9 & 35.6 &44.8  & 51.5 & 54.5   \\\hline
\textbf{R-C3D + In-Out}  &\textbf{12.6} & \textbf{23.0} & \textbf{32.1} & \textbf{41.1} & \textbf{49.2} & \textbf{53.9} & \textbf{56.2} \\
\textbf{R-C3D + Boundary} &\textbf{13.0} & \textbf{22.3} & \textbf{32.5} & \textbf{41.3} & \textbf{48.5} & \textbf{53.0} & \textbf{54.7}\\
\hline 
\end{tabular}
\label{tab:thumos-sota}
\end{table}

\textbf{ActivityNet \emph{v1.3}.} The comparison results on the ActivityNet \emph{v1.3} testing set are shown in Table \ref{tab:anet-sota}.  The results show that after using BLP models to refine temporal boundaries, we gain obvious improvement over the baseline R-C3D \cite{xu2017r} in terms of all range of tIoU and the average mAP. Meanwhile, compared with the state-of-the-art method CDC \cite{shou2017cdc}, our method shows competitive performance and get 15.5$\times$ relative gain when the tIoU is high (tIoU=0.95). This indicates that after the refinement, the segments have more precise boundaries and have larger overlap with ground truth instances.

\vspace{-1.3em}
%

\begin{table}[h]
\centering
\ninept
\caption{Temporal action detection results on ActivityNet \emph{v1.3} testing set (\%mAP@tIoU). We only list C3D feature based methods.}
\begin{tabular}{p{2.8cm}<{\raggedright}||p{0.8cm}<{\centering} p{0.8cm}<{\centering} p{0.8cm}<{\centering} || p{1.0cm}<{\centering} }
    \hline
Detection Method                   & 0.95            & 0.75           & 0.5          & Average \\
\hline
Wang \emph{et al.} \cite{upc2016} & 0.06		& 2.88				& 42.48 	& 14.62 \\
CDC \cite{shou2017cdc} & 0.20 & \textbf{25.70} &\textbf{43.00} & \textbf{22.90} \\ 
R-C3D (Baseline) \cite{xu2017r}   & 1.69         & 11.47          & 26.45          & 13.33          \\ \hline 
\textbf{R-C3D + In-Out} & \textbf{2.70} & 14.90 & 27.18 & 15.62 \\ 
\textbf{R-C3D + Boundary} & \textbf{3.30} & 16.31 & 28.10 & 16.78 \\ \hline
\end{tabular}

\label{tab:anet-sota}
\end{table}

\vspace{-1.5em}
\section{Conclusion}
\label{sec:conclusion}
In this paper, we propose a novel Boundary Likelihood Pinpointing (BLP) network for accurate temporal action localization. Specifically, instead of using boundary regression, we propose a substitution paradigm called boundary pinpointing. The localization process starts by assigning conditional probabilities to each equally divided unit of a search interval. These probabilities provide a measurement of confidence for each unit being within an action instance or being at the two boundaries. We can exploit these probabilities to accurately pinpoint the temporal boundaries under a simple probabilistic framework. Extensive experiments demonstrate that effectiveness of BLP localization model. Integrating our BLP model with existing action classifier into detection pipeline, competitive detection performance is achieved and we get 34.5\% (tIoU = 0.7) and 15.5$\times$ (tIoU = 0.95) relative gain over the mAP of state-of-the-art detectors on THUMOS'14 and ActivityNet respectively. 

 

 

\vfill\pagebreak
\bibliographystyle{IEEEbib}
\bibliography{strings,refs}

\begin{thebibliography}{10}

\bibitem{shou2016scnn}
Zheng Shou, Dongang Wang, and Shih-Fu Chang,
\newblock ``Temporal action localization in untrimmed videos via multi-stage
  cnns,''
\newblock in {\em Proceedings of the IEEE Conference on Computer Vision and
  Pattern Recognition}, 2016, pp. 1049--1058.

\bibitem{shou2017cdc}
Zheng Shou, Jonathan Chan, Alireza Zareian, Kazuyuki Miyazawa, and Shih-Fu
  Chang,
\newblock ``Cdc: convolutional-de-convolutional networks for precise temporal
  action localization in untrimmed videos,''
\newblock in {\em 2017 IEEE Conference on Computer Vision and Pattern
  Recognition (CVPR)}. IEEE, 2017, pp. 1417--1426.

\bibitem{zhao2017ssn}
Yue Zhao, Yuanjun Xiong, Limin Wang, Zhirong Wu, Xiaoou Tang, and Dahua Lin,
\newblock ``Temporal action detection with structured segment networks,''
\newblock in {\em The IEEE International Conference on Computer Vision (ICCV)},
  2017, vol.~8.

\bibitem{gao2017cascaded}
Jiyang Gao, Zhenheng Yang, and Ram Nevatia,
\newblock ``Cascaded boundary regression for temporal action detection,''
\newblock in {\em Proceedings of the British Machine Vision Conference (BMVC)},
  2017.

\bibitem{dai2017tcn}
Xiyang Dai, Bharat Singh, Guyue Zhang, Larry~S Davis, and Yan~Qiu Chen,
\newblock ``Temporal context network for activity localization in videos,''
\newblock in {\em 2017 IEEE International Conference on Computer Vision
  (ICCV)}. IEEE, 2017, pp. 5727--5736.

\bibitem{xu2017r}
Huijuan Xu, Abir Das, and Kate Saenko,
\newblock ``R-c3d: Region convolutional 3d network for temporal activity
  detection,''
\newblock in {\em The IEEE International Conference on Computer Vision (ICCV)},
  2017, vol.~6, p.~8.

\bibitem{lin2017sst}
Tianwei Lin, Xu~Zhao, and Zheng Shou,
\newblock ``Single shot temporal action detection,''
\newblock in {\em Proceedings of the 2017 ACM on Multimedia Conference}. ACM,
  2017, pp. 988--996.

\bibitem{sstad_buch_bmvc17}
Shyamal Buch, Victor Escorcia, Bernard Ghanem, Li~Fei-Fei, and Juan~Carlos
  Niebles,
\newblock ``End-to-end, single-stream temporal action detection in untrimmed
  videos,''
\newblock in {\em Proceedings of the British Machine Vision Conference
  ({BMVC})}, 2017.

\bibitem{Yang:2017wa}
Ke~Yang, Peng Qiao, Dongsheng Li, Shaohe Lv, and Yong Dou,
\newblock ``{Exploring Temporal Preservation Networks for Precise Temporal
  Action Localization},''
\newblock {\em arXiv.org}, Aug. 2017.

\bibitem{heilbron2017scc}
F~Caba Heilbron, Wayner Barrios, Victor Escorcia, and Bernard Ghanem,
\newblock ``Scc: Semantic context cascade for efficient action detection,''
\newblock in {\em IEEE Conference on Computer Vision and Pattern Recognition
  (CVPR)}, 2017, vol.~2.

\bibitem{Lea:2017fy}
Colin Lea, Michael~D Flynn, Rene Vidal, Austin Reiter, and Gregory~D Hager,
\newblock ``{Temporal Convolutional Networks for Action Segmentation and
  Detection},''
\newblock in {\em 2017 IEEE Conference on Computer Vision and Pattern
  Recognition (CVPR)}. 2017, pp. 1003--1012, IEEE.

\bibitem{alwassel_2018_detad}
Humam Alwassel, Fabian Caba~Heilbron, Victor Escorcia, and Bernard Ghanem,
\newblock ``Diagnosing error in temporal action detectors,''
\newblock in {\em The European Conference on Computer Vision (ECCV)},
  September.

\bibitem{lin2017temporal}
Tianwei Lin, Xu~Zhao, and Zheng Shou,
\newblock ``Temporal convolution based action proposal: Submission to
  activitynet 2017,''
\newblock {\em arXiv preprint arXiv:1707.06750}, 2017.

\bibitem{gao2017turn}
Jiyang Gao, Zhenheng Yang, Kan Chen, Chen Sun, and Ram Nevatia,
\newblock ``Turn tap: Temporal unit regression network for temporal action
  proposals,''
\newblock in {\em The IEEE International Conference on Computer Vision (ICCV)},
  Oct 2017.

\bibitem{chao2018rethinking}
Yu-Wei Chao, Sudheendra Vijayanarasimhan, Bryan Seybold, David~A Ross, Jia
  Deng, and Rahul Sukthankar,
\newblock ``Rethinking the faster r-cnn architecture for temporal action
  localization,''
\newblock in {\em Proceedings of the IEEE Conference on Computer Vision and
  Pattern Recognition}, 2018, pp. 1130--1139.

\bibitem{THUMOS14}
Y.-G. Jiang, J.~Liu, A.~Roshan~Zamir, G.~Toderici, I.~Laptev, M.~Shah, and
  R.~Sukthankar,
\newblock ``{THUMOS} challenge: Action recognition with a large number of
  classes,'' \url{http://crcv.ucf.edu/THUMOS14/}, 2014.

\bibitem{caba2015activitynet}
Bernard~Ghanem Fabian Caba~Heilbron, Victor~Escorcia and Juan~Carlos Niebles,
\newblock ``Activitynet: A large-scale video benchmark for human activity
  understanding,''
\newblock in {\em Proceedings of the IEEE Conference on Computer Vision and
  Pattern Recognition}, 2015, pp. 961--970.

\bibitem{simonyan2014two}
Karen Simonyan and Andrew Zisserman,
\newblock ``Two-stream convolutional networks for action recognition in
  videos,''
\newblock in {\em Advances in neural information processing systems}, 2014, pp.
  568--576.

\bibitem{tran2015learning}
Du~Tran, Lubomir Bourdev, Rob Fergus, Lorenzo Torresani, and Manohar Paluri,
\newblock ``Learning spatiotemporal features with 3d convolutional networks,''
\newblock in {\em Computer Vision (ICCV), 2015 IEEE International Conference
  on}. IEEE, 2015, pp. 4489--4497.

\bibitem{Wang:2016fa}
Limin Wang, Yuanjun Xiong, Zhe Wang, Yu~Qiao, Dahua Lin, Xiaoou Tang, and Luc
  Van~Gool,
\newblock ``Temporal segment networks: Towards good practices for deep action
  recognition,''
\newblock in {\em European Conference on Computer Vision}. Springer, 2016, pp.
  20--36.

\bibitem{carreira2017quo}
Joao Carreira and Andrew Zisserman,
\newblock ``Quo vadis, action recognition? a new model and the kinetics
  dataset,''
\newblock in {\em 2017 IEEE Conference on Computer Vision and Pattern
  Recognition (CVPR)}. IEEE, 2017, pp. 4724--4733.

\bibitem{wu2018compressed}
Chao-Yuan Wu, Manzil Zaheer, Hexiang Hu, R~Manmatha, Alexander~J Smola, and
  Philipp Kr{\"a}henb{\"u}hl,
\newblock ``Compressed video action recognition,''
\newblock in {\em Proceedings of the IEEE Conference on Computer Vision and
  Pattern Recognition}, 2018, pp. 6026--6035.

\bibitem{gidaris2016locnet}
Spyros Gidaris and Nikos Komodakis,
\newblock ``Locnet: Improving localization accuracy for object detection,''
\newblock in {\em Proceedings of the IEEE Conference on Computer Vision and
  Pattern Recognition}, 2016, pp. 789--798.

\bibitem{BSN2018arXiv}
Tianwei Lin, Xu~Zhao, Haisheng Su, Chongjing Wang, and Ming Yang,
\newblock ``Bsn: Boundary sensitive network for temporal action proposal
  generation,''
\newblock in {\em European Conference on Computer Vision}, 2018.

\bibitem{escorcia2016daps}
Victor Escorcia, Fabian~Caba Heilbron, Juan~Carlos Niebles, and Bernard Ghanem,
\newblock ``Daps: Deep action proposals for action understanding,''
\newblock in {\em European Conference on Computer Vision}. Springer, 2016, pp.
  768--784.

\bibitem{softnms}
Navaneeth Bodla, Bharat Singh, Rama Chellappa, and Larry~S. Davis,
\newblock ``Soft-nms -- improving object detection with one line of code,''
\newblock 2017.

\bibitem{jia2014caffe}
Yangqing Jia, Evan Shelhamer, Jeff Donahue, Sergey Karayev, Jonathan Long, Ross
  Girshick, Sergio Guadarrama, and Trevor Darrell,
\newblock ``Caffe: Convolutional architecture for fast feature embedding,''
\newblock in {\em Proceedings of the 22nd ACM international conference on
  Multimedia}. ACM, 2014, pp. 675--678.

\bibitem{upc2016}
R.~Wang and D.~Tao,
\newblock ``Uts at activitynet 2016,''
\newblock {\em AcitivityNet Large Scale Activity Recognition Challenge}, 2016.

\end{thebibliography}

\end{document}